\documentclass{article}
\usepackage{nips07submit_e,times}
\usepackage[dvips]{graphicx}
\usepackage{bm}
\usepackage{amssymb,amsfonts,amsmath}

\newtheorem{definition}{Definition}
\newtheorem{theorem}{Theorem}

\long\def\symbolfootnote[#1]#2{\begingroup%
\def\thefootnote{\fnsymbol{footnote}}\footnote[#1]{#2}\endgroup}

\title{
\mbox{Simulated Annealing: Rigorous finite-time guarantees}\\
for optimization on continuous domains$^*$
}

\author{
Andrea Lecchini-Visintini\\
Department of Engineering\\
University of Leicester, UK \\
\texttt{alv1@leicester.ac.uk}
\And
John Lygeros\\
Automatic Control Laboratory\\
ETH Zurich, Switzerland.\\
\texttt{lygeros@control.ee.ethz.ch}
\And
Jan Maciejowski \\
Department of Engineering \\
University of Cambridge, UK \\
\texttt{jmm@eng.cam.ac.uk}
}

\begin{document}

\maketitle

\symbolfootnote[0]{$^*${\bf Preprint.} The final version will appear in: Advances in Neural Information Processing Systems 20, Proceedings of NIPS 2007, MIT Press.}

\begin{abstract}
Simulated annealing is a popular method for approaching the solution
of a global optimization problem. Existing results on its
performance apply to discrete combinatorial optimization where the
optimization variables can assume only a finite set of possible
values. We introduce a new general formulation of simulated
annealing which allows one to guarantee finite-time performance in
the optimization of functions of continuous variables. The results
hold universally for {\it any} optimization problem on a bounded
domain and establish a connection between simulated annealing and
up-to-date theory of convergence of Markov chain Monte Carlo methods
on continuous domains. This work is inspired by the concept of
finite-time learning with known accuracy and confidence developed in
statistical learning theory.
\end{abstract}

Optimization is the general problem of finding a value of a vector
of variables $\theta$ that maximizes (or minimizes) some scalar
criterion $U(\theta)$. The set of all possible values of the vector
$\theta$ is called the optimization domain. The elements of $\theta$
can be discrete or continuous variables. In the first case the
optimization domain is usually finite, such as in the well-known
traveling salesman problem; in the second case the optimization
domain is a continuous set. An important example of a continuous
optimization domain is the set of 3-D configurations of a sequence
of amino-acids in the problem of finding the minimum energy folding
of the corresponding protein \cite{Wales-03}.

In principle, any optimization problem on a finite domain can be
solved by an exhaustive search. However, this is often beyond
computational capacity: the optimization domain of the traveling
salesman problem with $100$ cities contains more than $10^{155}$
possible tours. An efficient algorithm to solve the traveling
salesman and many similar problems has not yet been found and such
problems remain reliably solvable only in principle
\cite{Achlioptas-et-al-05}. Statistical mechanics has inspired
widely used methods for finding good approximate solutions in hard
discrete optimization problems which defy efficient exact solutions
\cite{Kirkpatrick-et-al-83,Bonomi-Lutton-84,Fu-Anderson-86,Mezard-et-al-02}.
Here a key idea has been that of simulated annealing
\cite{Kirkpatrick-et-al-83}: a random search based on the
Metropolis-Hastings algorithm, such that the distribution of the
elements of the domain visited during the search converges to an
equilibrium distribution concentrated around the global optimizers.
Convergence and finite-time performance of simulated annealing on
finite domains has been evaluated in many works,
e.g. \cite{Laarhoven-Aarts-87,Mitra-et-al-86,Hajek-88,Hannig-et-al-06}.

On continuous domains, most popular optimization methods perform a
local gradient-based search and in general converge to local
optimizers; with the notable exception of convex criteria where
convergence to the unique global optimizer  occurs
\cite{Boyd-Vandenberghe-03}. Simulated annealing performs a global
search and can be easily implemented on continuous domains. Hence it
can be considered a powerful complement to local methods. In this
paper,  we introduce  for the first time rigorous guarantees on the
finite-time performance of simulated annealing on continuous
domains. We will show that it is possible to derive simulated
annealing algorithms  which, with an arbitrarily high level of
confidence, find an approximate solution to the problem of
optimizing a function of continuous variables, within a specified
tolerance to the global optimal solution after a known finite number
of steps. Rigorous guarantees on the finite-time performance of
simulated annealing in the optimization of functions of continuous
variables have never been obtained before; the only results
available state that simulated annealing converges to a global
optimizer as the number of steps grows to infinity,
e.g.~\cite{Haario-Saksman-91,Gelfand-Mitter-91,Tsallis-Stariolo-96,Locatelli-00}.\medskip\linebreak
The background of our work is twofold. On the one hand, our notion
of  approximate solution to a global optimization problem is
inspired by the concept of finite-time learning with known accuracy
and confidence developed in statistical learning theory
\cite{Vapnik-95,Vidyasagar-03}. We actually maintain an important
aspect of statistical learning theory which is that we do not
introduce any particular assumption on the optimization criterion,
i.e.~our results hold regardless of what $U$ is. On the other hand,
we ground our results on the theory of convergence, with
quantitative bounds on the distance to the target distribution, of
the Metropolis-Hastings algorithm and Markov Chain Monte Carlo
(MCMC) methods, which has been one of the main achievements of
recent research in statistics
\cite{Meyn-Tweedie-93,Rosenthal-95,Mengersen-Tweedie-96,Roberts-Rosenthal-04}.

In this paper, we will not develop any  ready-to-use optimization
algorithm. We will instead introduce a general formulation of the
simulated annealing method which allows one to derive new simulated
annealing algorithms with rigorous finite-time guarantees on the
basis of existing theory. The Metropolis-Hastings algorithm and the
general family of MCMC  methods have many degrees of freedom. The
choice and comparison of specific algorithms goes beyond the scope
of the paper.

The paper is organized in the following sections. In {\it Simulated
annealing} we introduce the method and fix the notation. In {\it
Convergence} we recall the reasons why finite-time guarantees for
simulated annealing on continuous domains have not been obtained
before. In {\it Finite-time guarantees} we present the main result
of the paper.  In {\it Conclusions} we state our
findings and conclude the paper.\vspace{-0.05cm}

\section{Simulated annealing}
The original formulation of simulated annealing was inspired by the
analogy between the stochastic evolution of the thermodynamic state
of an annealing material towards the  configurations of  minimal
energy and the search for the global minimum of an optimization
criterion \cite{Kirkpatrick-et-al-83}. In the procedure, the
optimization criterion plays the role of the energy and the state of
the annealed material is simulated by the evolution of the state of
an inhomogeneous Markov chain. The state of the chain evolves
according to the Metropolis-Hastings algorithm in order to simulate
the Boltzmann distribution of thermodynamic equilibrium. The
Boltzmann distribution is simulated for a decreasing sequence of
temperatures (``cooling"). The target distribution of the cooling
procedure is the limiting Boltzmann distribution, for the
temperature that tends to zero, which takes non-zero values only on
the set of global minimizers \cite{Laarhoven-Aarts-87}.

The original formulation of the method was for a finite domain.
However, simulated annealing can be generalized straightforwardly to
a continuous domain because the Metropolis-Hastings algorithm can be
used with almost no differences on discrete and continuous domains
The main difference is that  on a continuous domain the equilibrium
distributions are specified by probability densities. On a
continuous domain, Markov transition kernels in which the
distribution of the elements visited by the chain converges to an
equilibrium distribution with the desired density can be constructed
using the Metropolis-Hastings algorithm and the general family of
MCMC methods \cite{Robert-Casella-04}.

We point out that Boltzmann distributions are not the only
distributions which can be adopted as equilibrium distributions in
simulated annealing \cite{Laarhoven-Aarts-87}. In this paper it is
convenient for us to adopt a different type of equilibrium
distribution in place of Boltzmann distributions.

\subsection{Our setting}
The optimization criterion is $U:\bm{\Theta}\rightarrow [0,\, 1]$,
with $\bm{\Theta}\subset\mathbb{R}^N$. The  assumption  that $U$
takes values in the interval $[0,\, 1]$ is a technical one. It does
not imply any serious loss of generality. In general, any bounded
optimization criterion can be scaled to take values in $[0,\, 1]$.
We assume that the optimization task is to find a global maximizer;
this can be done without loss of generality. We also assume that
$\bm{\Theta}$ is a bounded set.

\begin{figure}[t]
\includegraphics[width=\columnwidth]{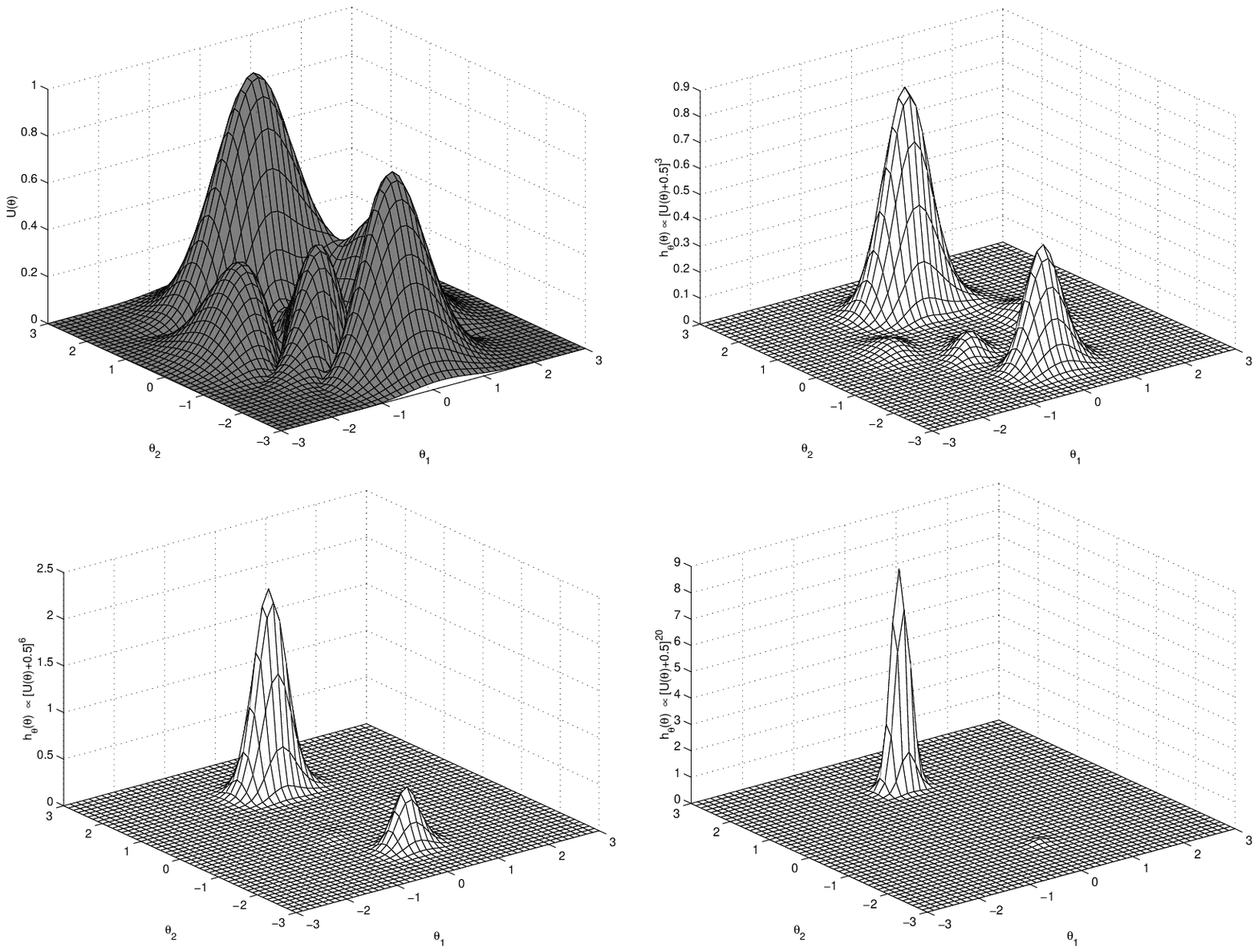}
\caption{The function $U(\theta)$ (upper left) and some probability
densities of the form $h_{\bm{\theta}}(\theta)\propto [U(\theta)+
\delta]^{J}$ for $\delta=0.5$ and $J=3$ (upper right), $J=6$ (lower
left) and $J=20$ (lower right).}\label{fig:figfun}
\end{figure}

We consider equilibrium distributions defined by probability density
functions proportional to $[U(\theta)+\delta]^J$ where $J$ and
$\delta$ are two strictly positive parameters. We use $\pi^{(J)}$ to
denote an equilibrium distribution,
i.e.~$\pi^{(J)}(d\theta)\propto[U(\theta)+\delta]^J\pi_{Leb}(d\theta)$
where $\pi_{Leb}$ is the standard Lebesgue measure.
Here, $J^{-1}$ plays the role of the temperature: if the function
$U(\theta)$ (Figure \ref{fig:figfun}.a) plus $\delta$ is taken to a positive power
$J$ then as $J$ increases (i.e. as $J^{-1}$ decreases)
$[U(\theta)+\delta]^J$  (Figure \ref{fig:figfun}.b-d) becomes increasingly peaked
around the global maximizers.
The parameter  $\delta$ is an  offset which guarantees
that the equilibrium densities are always strictly positive, even if
$U$ takes zero values on some elements of the domain. The offset
$\delta$ is chosen by the user and we show later that our results
allow one to make an optimal selection of $\delta$. The
zero-temperature  distribution is the limiting distribution, for
$J\rightarrow \infty$, which takes non-zero values only on the set
of global maximizers. It  is denoted by $\pi^{(\infty)}$.\medskip\linebreak
In the generic formulation of the method, the Markov transition
kernel of the $k$-th step of the inhomogeneous chain has equilibrium
distribution $\pi^{(J_k)}$ where $\{J_k\}_{k=1,2,\dots}$ is the
``cooling schedule". The cooling schedule is a non-decreasing
sequence of positive numbers according to which the equilibrium
distribution become increasingly sharpened  during the evolution of
the chain. We use $\bm{\theta}_k$ to denote the state of the chain
and $P_{\bm{\theta}_k}$ to denote its probability distribution. The
distribution $P_{\bm{\theta}_k}$ obviously depends on the initial
condition $\bm{\theta}_0$. However, in this work, we don't need to
make this dependence explicit in the notation.

{\it Remark 1:} If, given an element $\theta$ in $\bm{\Theta}$, the
value $U(\theta)$ can be computed directly, we say that $U$ is a
deterministic criterion, e.g.~the energy landscape in protein
structure prediction \cite{Wales-03}. In problems involving random
variables, the value $U(\theta)$ may be the expected value
$U(\theta) = \int g(x,\theta)p_{\bm{x}}(x;\theta)dx$ of some
function $g$ which depends on both the optimization variable
$\theta$, and on some random variable $\bm{x}$ which has probability
density $p_{\bm{x}}(x;\theta)$ (which may itself depend on
$\theta$). In such problems it is usually not possible to compute
$U(\theta)$ directly, either because evaluation of the integral
requires too much computation, or because no analytical expression
for $p_{\bm{x}}(x;\theta)$ is available. Typically one must perform
stochastic simulations in order to obtain samples of $\bm{x}$ for a
given $\theta$, hence obtain sample values of $g(\bm{x},\theta)$,
and thus construct a Monte Carlo estimate of $U(\theta)$. The
Bayesian design of clinical trials is an important application area
where such expected-value criteria arise
\cite{Spiegelhalter-et-al-2004}.
The authors of this paper investigate the optimization of expected-value criteria
motivated by problems of aircraft routing \cite{Leccchini-et-al-2006}.
In the particular case that $p_{\bm{x}}(x;\theta)$ does not depend on
$\theta$, the optimization task is often called ``empirical risk
minimization'', and is studied extensively in statistical learning
theory \cite{Vapnik-95,Vidyasagar-03}. The results of this paper
apply in the same way to the optimization of both deterministic and
expected-value criteria. The MCMC method developed by M\"{u}ller
\cite{Muller-99,Muller-et-al-04} allows one to construct simulated
annealing algorithms for the optimization of expected-value
criteria. M\"{u}ller  \cite{Muller-99,Muller-et-al-04} employs the
same equilibrium distributions as those described in our setting; in
his context $J$ is restricted to integer values.

\begin{figure}[t]\centering
\includegraphics[width=0.9\columnwidth]{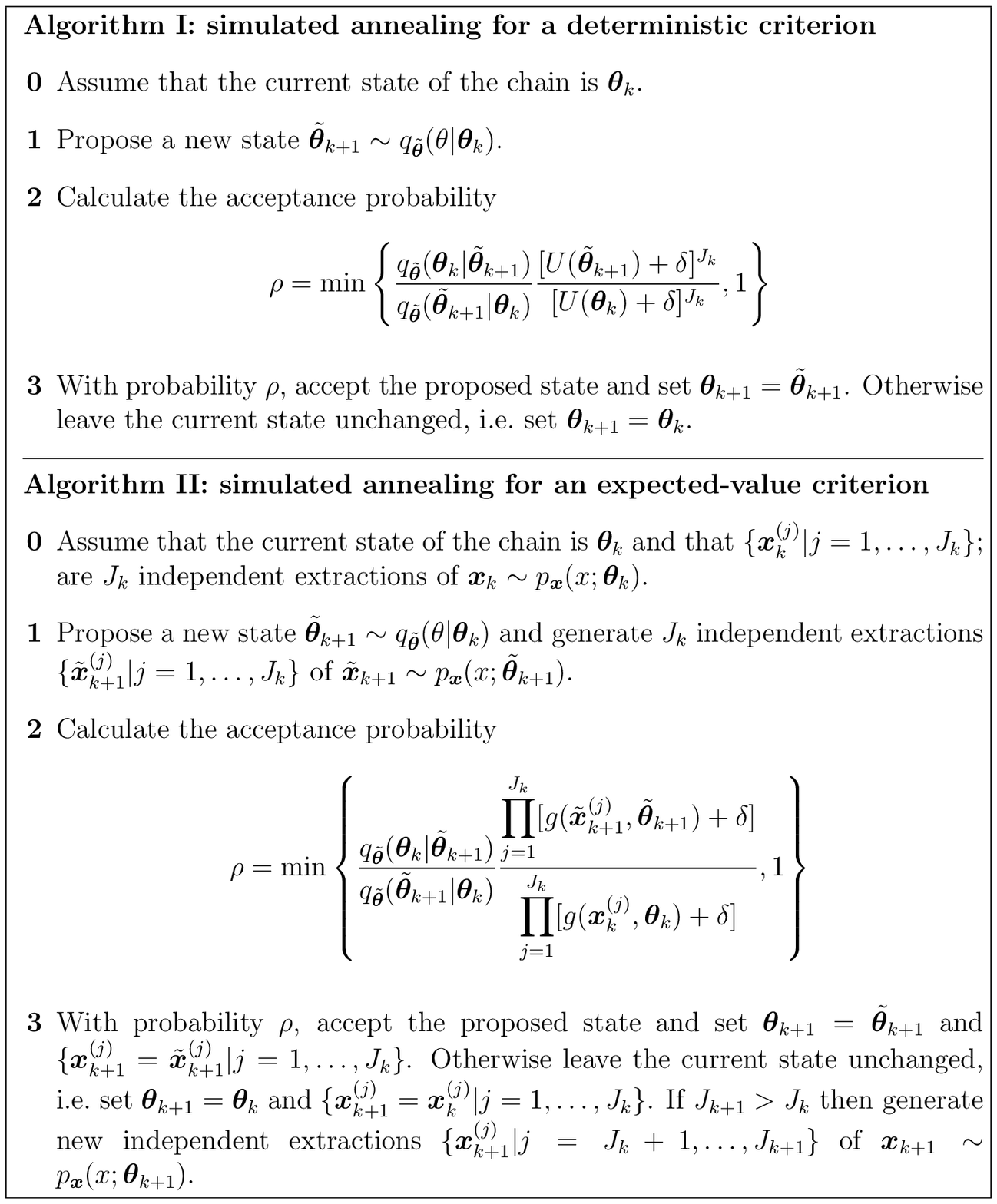}
\caption{The basic iterations of simulated annealing with
equilibrium distributions $\pi^{(J)}(d\theta)$, for the maximization
of deterministic and expected-value criteria (see Remark 1). In the
algorithms, $q_{\tilde{\bm{\theta}}}$ is the density of the
``proposal distribution'' of the Metropolis step. The iteration for
the expected value criterion has been proposed by M\"{u}ller
\cite{Muller-99,Muller-et-al-04}. }\label{fig:algo}\vspace{-1cm}
\end{figure}

In Figure \ref{fig:algo}, we illustrate the basic iteration of a generic
simulated annealing algorithm with equilibrium distributions
$\pi^{(J)}(d\theta)$ for the optimization of deterministic
and expected-value criteria.

\section{Convergence}

The rationale of simulated annealing is as follows: if the
temperature is kept constant, say $J_k=J$, then the distribution of
the state of the chain  $P_{\bm{\theta}_k}$ tends to the equilibrium
distribution $\pi^{(J)}$; if $J\rightarrow\infty$ then the
equilibrium distribution $\pi^{(J)}$ tends to the zero-temperature
distribution $\pi^{(\infty)}$; as a result, if the cooling schedule
$J_k$ tends to infinity, one obtains that $P_{\bm{\theta}_k}$
``follows'' $\pi^{(J_k)}$ and that  $\pi^{(J_k)}$ tends to
$\pi^{(\infty)}$ and eventually that the distribution of the state
of the chain  $P_{\bm{\theta}_k}$ tends to $\pi^{(\infty)}$. The
theory shows that, under conditions on the cooling schedule and the
Markov transition kernels, the distribution of the state of the
chain $P_{\bm{\theta}_k}$ actually converges to the target
zero-temperature distribution $\pi^{(\infty)}$ as
$k\rightarrow\infty$
\cite{Haario-Saksman-91,Gelfand-Mitter-91,Tsallis-Stariolo-96,Locatelli-00}.
Convergence to the zero-temperature distribution implies that
asymptotically the state of the chain eventually coincides with a
global optimizer with probability one.

The difficulty which must be overcome in order to obtain finite step
results on simulated annealing algorithms on a continuous domain is
that usually, in an optimization problem defined over continuous
variables, the set of global optimizers has zero Lebesgue measure
(e.g.~a set of isolated points). If the set of global optimizers has
zero measure then the set of global optimizers has null probability
according to the equilibrium distributions  $\pi^{(J)}$ for any
finite $J$ and, as a consequence, according to the distributions
 $P_{\bm{\theta}_k}$ for any finite  $k$.
Put another way, the probability that the state of the chain visits
the set of global optimizers is constantly zero after any finite
number of steps. Hence  the confidence of the fact that the solution
provided by the algorithm in finite time coincides with a global
optimizer is also constantly zero. Notice that this is not the case
for a finite domain, where the set of global optimizers is of
non-null measure with respect to the reference counting measure
\cite{Laarhoven-Aarts-87,Mitra-et-al-86,Hajek-88,Hannig-et-al-06}.

It is instructive to look at the issue also in terms of the rate of
convergence to the target zero-temperature distribution. On a
discrete domain, the distribution of the state of the chain at each
step and the zero-temperature distribution are both standard
discrete distributions. It is then possible to define a distance
between them and study the rate of convergence of this distance to
zero. This analysis allows one to obtain results on the finite-time
behavior of simulated annealing
\cite{Laarhoven-Aarts-87,Mitra-et-al-86}. On a continuous domain and
for a set of global optimizers of measure zero, the target
zero-temperature distribution $\pi^{(\infty)}$ ends up being a
mixture of probability masses on the set of global optimizers. In
this situation, although the distribution of the state of the chain
$P_{\bm{\theta}_k}$ still converges asymptotically  to
$\pi^{(\infty)}$, it is not possible to introduce a sensible
distance between the two distributions and a rate of convergence to
the target distribution cannot even be defined (weak convergence),
see \cite[Theorem 3.3]{Haario-Saksman-91}. This is the reason that
until now there have been no guarantees on the performance of
simulated annealing on a continuous domain after a finite number of
computations: by adopting the zero-temperature distribution
$\pi^{(\infty)}$ as the target distribution it is only possible to
prove asymptotic convergence in infinite time to a global optimizer.

{\it Remark 2:} The standard distance between two distributions, say
$\mu_1$ and $\mu_2$, on a continuous support is the total variation
norm $\|\mu_1 - \mu_2\|_{TV} = \sup_{A} | \mu_1(A) - \mu_2(A)|$, see
e.g.~\cite{Roberts-Rosenthal-04}. In simulated annealing on a
continuous domain the distribution of the state of the chain
$P_{\bm{\theta}_k}$ is absolutely continuous with respect to the
Lebesgue measure (i.e.~\mbox{$\pi_{Leb}(A)=0\Rightarrow
P_{\bm{\theta}_k}(A)=0$}), by construction for any finite $k$. Hence
if the set of global optimizers has zero Lebesgue measure then it has
zero measure also according to $P_{\bm{\theta}_k}$. The set of
global optimizers has however measure 1  according to
$\pi^{(\infty)}$. The distance $\|P_{\bm{\theta}_k} -
\pi^{(\infty)}\|_{TV}$ is then constantly $1$ for any finite $k$.

It is also worth mentioning that if the set of global optimizers has
zero measure then asymptotic convergence to the zero-temperature
distribution $\pi^{(\infty)}$ can be proven only under the
additional  assumptions of continuity and differentiability of $U$
\cite{Haario-Saksman-91,Gelfand-Mitter-91,Tsallis-Stariolo-96,Locatelli-00}.

\section{Finite-time guarantees}

In general, optimization algorithms for problems defined on
continuous variables
 can only find  approximate solutions in finite time \cite{Blum-et-al-98}.
Given an element $\theta$ of a continuous domain how can we assess
how good it is as an approximate solution to an optimization
problem? Here we introduce the concept of {\it approximate global
optimizer} to answer this question. The definition is given for a
maximization problem in a continuous but bounded domain. We use two
parameters: the {\it value imprecision} $\epsilon$ (greater than or
equal to 0) and the {\it residual domain} $\alpha$ (between 0  and
1) which together determine the level of approximation. We say that
$\theta$ is an approximate global optimizer of $U$ with value
imprecision $\epsilon$ and residual domain $\alpha$ if the function
$U$ takes values strictly greater than $U(\theta) + \epsilon$ only
on a subset of values of $\theta$ no larger than an $\alpha$ portion
of the optimization domain. The formal definition is as follows.

\begin{definition}
Let $U:\bm{\Theta}\rightarrow\mathbb{R}$ be an  optimization
criterion where $\bm{\Theta}\subset\mathbb{R}^N$ is bounded. Let
$\pi_{Leb}$ denote the standard Lebesgue measure. Let $\epsilon\geq
0$ and  $\alpha\in [0,\,1]$ be given numbers. Then $\theta$ is an
{\it approximate global optimizer} of $U$ with value imprecision
$\epsilon$ and residual domain $\alpha$ if
$\pi_{Leb}\{\theta'\in\bm{\Theta} :  U(\theta') > U(\theta) +
\epsilon \} \leq \alpha\, \pi_{Leb}(\bm{\Theta})\, .$
\end{definition}

In other words, the value $U(\theta)$ is within $\epsilon$ of a
value which is greater than the values that $U$ takes on at least a
$1-\alpha$ portion of the domain. The smaller $\epsilon$  and
$\alpha$ are, the better is the approximation of a true global
optimizer. If  both $\alpha$ and $\epsilon$ are equal to zero then
$U(\theta)$ coincides with the  essential supremum of $U$.

Our definition of approximate global optimizer carries an important
property, which holds regardless of what the criterion $U$ is: if
$\epsilon$ and $\alpha$ have non-zero values then the set of
approximate global optimizers always has non-zero Lebesgue measure.
It follows that the probability that the chain visits the set of
approximate global optimizers can be non-zero. Hence, it is sensible
to study the confidence of the fact that the solution found by
simulated annealing in finite time is an approximate global
optimizer.

{\it Remark 3:} The intuition that our notion of approximate global
optimizer can be used to obtain formal guarantees on the finite-time
performance of optimization methods based on a stochastic search of
the domain is already apparent in the work of Vidyasagar
\cite{Vidyasagar-03,Vidyasagar-01}. Vidyasagar
\cite{Vidyasagar-03,Vidyasagar-01} introduces a similar definition
and obtains rigorous finite-time guarantees in the optimization of
expected value criteria based on uniform independent sampling of the
domain. Notably, the number of independent samples required to
guarantee some desired accuracy and confidence turns out to be
polynomial in the values of the desired imprecision, residual domain
and confidence. Although the method of Vidyasagar  is not highly
sophisticated, it has had considerable success in solving difficult
control system design applications
\cite{Vidyasagar-01,Tempo-et-al-05}. Its  appeal stems from its
rigorous finite-time guarantees which exist without the need for any
particular  assumption on the optimization criterion.

Here we show that finite-time guarantees for simulated annealing can
be obtained by selecting a distribution $\pi^{(J)}$ with a finite
$J$ as the target distribution in place of the zero-temperature
distribution $\pi^{(\infty)}$. The fundamental result is the
following theorem which allows one to select in a rigorous way
$\delta$ and $J$ in the target distribution $\pi^{(J)}$. It is
important to stress that the result holds universally for {\it any}
optimization criterion $U$ on a bounded domain. The only  minor
requirement is that $U$ takes values in $[0,\, 1]$.

\begin{theorem}\label{th:confidence}
Let $U:\bm{\Theta}\rightarrow[0,\,1]$ be an optimization criterion
where $\bm{\Theta}\subset\mathbb{R}^N$ is bounded. Let $J\geq 1$ and
$\delta>0$ be given numbers. Let $\bm{\theta}$ be a multivariate
random variable with distribution $\pi^{(J)}(d\theta)\propto
[U(\theta)+\delta]^J\pi_{Leb}(d\theta)$. Let $\alpha\in (0,\, 1]$
and $\epsilon\in [0,\, 1]$ be given numbers and define
\begin{equation}\label{eq:confidence}
\sigma = \frac { 1 } { \mbox{$\displaystyle 1 + \left[\frac{1 +
\delta}{\epsilon+1+\delta}\right]^{\,J}
\left[\frac{1}{\alpha}\frac{1 +\delta}{\epsilon+\delta} - 1\right]
\frac{1+\delta}{\delta} $}}\,\, .
\end{equation}
 Then the statement
``$\bm{\theta}$ is an approximate global optimizer of $U$ with value
imprecision  $\epsilon$ and residual domain $\alpha$" holds with
probability at least $\sigma$.
\end{theorem}
{\it Proof.} See Appendix A.

The importance of the choice of a target distribution $\pi^{(J)}$
with a finite $J$ is that $\pi^{(J)}$ is absolutely continuous with
respect to the Lebesgue measure. Hence, the distance
$\|P_{\bm{\theta}_k}  - \pi^{(J)}\|_{\mbox{\tiny TV}}$ between the
distribution of the state of the chain $P_{\bm{\theta}_k}$ and the
target distribution $\pi^{(J)}$ is a meaningful quantity.

Convergence of the Metropolis-Hastings algorithm and MCMC methods in
total variation norm is a well studied problem. The theory provides
simple conditions under which one derives upper bounds on the
distance to the target distribution which are known at each step of
the chain and decrease monotonically to zero as the number of steps
of the chain grows. The theory has been developed mainly for {\it
homogeneous} chains
\cite{Meyn-Tweedie-93,Rosenthal-95,Mengersen-Tweedie-96,Roberts-Rosenthal-04}.

In the case of simulated annealing, the factor that enables us to
employ these results is the absolute continuity of the target
distribution $\pi^{(J)}$ with respect to the Lebesgue measure.
However, simulated annealing involves the simulation of
inhomogeneous chains. In this respect, another important fact is
that the choice of a target distribution $\pi^{(J)}$ with a finite
$J$ implies that the inhomogeneous Markov chain can in fact be
formed by a finite sequence of homogeneous chains (i.e.~the cooling
schedule $\{J_k\}_{k=1,2,\dots}$ can be chosen to be a sequence that
takes only a finite set of values). In turn, this allows one to
apply the theory of homogeneous MCMC methods to study the
convergence of $P_{\bm{\theta}_k}$ to $\pi^{(J)}$ in total variation
norm.

On a bounded domain, simple conditions on the `proposal distribution'
in the  iteration of the simulated annealing algorithm  allows  one
to obtain upper bounds on $\|P_{\bm{\theta}_k}  - \pi^{(J)}\|_{\mbox{\tiny TV}}$ that decrease
geometrically to zero as $k\rightarrow\infty$, without the need for
any additional assumption on $U$ \cite{Meyn-Tweedie-93,Rosenthal-95,Mengersen-Tweedie-96,Roberts-Rosenthal-04}.

It is then appropriate to introduce the following finite-time result.
\begin{theorem}\label{th:confidence_k}
Let the notation and assumptions of Theorem \ref{th:confidence}
hold. Let $\bm{\theta}_k$, with distribution $P_{\bm{\theta}_k}$, be
the state of the inhomogeneous chain of a simulated annealing
algorithm with target distribution $\pi^{(J)}$. Then the statement
``$\bm{\theta}_k$ is an approximate global optimizer of $U$ with
value imprecision  $\epsilon$ and residual domain $\alpha$" holds
with probability at least $\sigma-\|P_{\bm{\theta}_k}  -
\pi^{(J)}\|_{\mbox{\tiny TV}}$.
\end{theorem}
The proof of the theorem follows directly from the definition of the
total variation norm.

It follows  that if simulated annealing is implemented with an
algorithm which converges in total variation distance to a target
distribution $\pi^{(J)}$ with a finite $J$, then one can  state
with confidence arbitrarily close to 1 that the solution found by
the algorithm after the known appropriate finite number of steps is
an approximate global optimizer with the desired approximation
level. For given non-zero values of $\epsilon$, $\alpha$ the value
of $\sigma$ given by (\ref{eq:confidence}) can be made arbitrarily
close to 1 by choice of $J$; while
 the distance $\|P_{\bm{\theta}_k}  - \pi^{(J)}\|_{\mbox{\tiny TV}}$
can be made arbitrarily small by taking the known sufficient number
of steps.

It can be shown that there exists the possibility of
making an optimal choice of $\delta$ and $J$ in the target distribution $\pi^{(J)}$.
In fact, for given $\epsilon$ and $\alpha$ and a given value of $J$ there exists
an optimal choice of $\delta$ which maximizes the value of
$\sigma$ given by (\ref{eq:confidence}).
Hence, it is possible to obtain a desired $\sigma$ with
the smallest possible $J$. The advantage of choosing  the smallest $J$,
consistent with the required approximation and confidence,
is that it will decrease the number of steps required to achieve
the desired reduction of $\|P_{\bm{\theta}_k}  - \pi^{(J)}\|_{\mbox{\tiny TV}}$.

\section{Conclusions}

We have introduced a new formulation of simulated annealing which
admits rigorous finite-time guarantees in the optimization of
functions of continuous variables. First, we have introduced the
notion of approximate global optimizer. Then, we have shown that
simulated annealing is guaranteed to find approximate global
optimizers, with the desired confidence and the desired level of
accuracy, in a known finite number of steps, if a proper choice of
the target distribution is made and conditions for convergence in
total variation norm are met. The results  hold for {\it any}
optimization criterion on a bounded domain with the only minor
requirement that it takes values between 0 and 1.

In this framework, simulated annealing algorithms with rigorous
finite-time guarantees can be derived by studying the choice of the
proposal distribution and of the cooling schedule, in the generic
iteration of simulated annealing, in order to ensure convergence to
the target distribution in total variation norm. To do this,
existing theory of convergence of the Metropolis-Hastings algorithm
and MCMC methods on continuous domains can be used
\cite{Meyn-Tweedie-93,Rosenthal-95,Mengersen-Tweedie-96,Roberts-Rosenthal-04}.

Vidyasagar \cite{Vidyasagar-03,Vidyasagar-01} has introduced a
similar definition of approximate global optimizer and has shown
that approximate optimizers with desired accuracy and confidence can
be obtained with a number of uniform independent samples of the
domain which is polynomial in the  accuracy and confidence
parameters. In general, algorithms developed with the MCMC
methodology can be expected to be equally or more efficient than
uniform independent sampling.

\subsubsection*{Acknowledgments}

Work supported by EPSRC, Grant EP/C014006/1, and by the European Commission under
projects HYGEIA FP6-NEST-4995 and iFly FP6-TREN-037180. We thank  S.~Brooks, M.~Vidyasagar
and D.~M.~Wolpert for discussions and useful comments on the paper.

\appendix

\section{Proof of Theorem \ref{th:confidence}}

Let $\bar\alpha\in (0,\, 1]$ and $\rho\in (0,\, 1]$ be given
numbers. Let $U_\delta(\theta):=U(\theta)+\delta$. Let  $\pi_\delta$
be a  normalized measure such that $\pi_\delta(d\theta) \propto
U_\delta(\theta)\pi_{Leb}(d\theta)$. In the first part of the proof
we find a lower bound on the probability that $\bm{\theta}$ belongs
to the set
$$
\{\theta\in\bm{\Theta} :
\pi_\delta\{\theta'\in\bm{\Theta}:\rho\,U_\delta(\theta') >
U_\delta(\theta)\} \leq \bar\alpha\}\, .
$$
Let $y_{\bar\alpha} := \inf \{y: \pi_\delta \{\theta\in\bm{\Theta}:
U_\delta(\theta)\leq y\} \geq  1-\bar\alpha \}$. To start with we
show that the set $\{\theta\in\bm{\Theta}:
\pi_\delta\{\theta'\in\bm{\Theta}: \rho\, U_\delta(\theta')>
U_\delta(\theta) \}\leq\bar\alpha \}$ coincides with $\{
\theta\in\bm{\Theta}: U_\delta(\theta) \geq \rho\, y_{\bar\alpha}
\}$.
 Notice that the quantity
$\pi_\delta \{\theta\in\bm{\Theta}:  U_\delta(\theta)\leq y\}$ is a
right-continuous non-decreasing function of $y$ because it has  the
form of a distribution function (see e.g.~\cite[p.162]{Gnedenko-68}
and \cite[Lemma 11.1]{Vidyasagar-03}). Therefore we have $\pi_\delta
\{\theta\in\bm{\Theta}: U_\delta(\theta)\leq y_{\bar\alpha}\} \geq
1-\bar\alpha$ and
$$
y\geq \rho\, y_{\bar\alpha}\quad\Rightarrow\quad \pi_\delta
\{\theta'\in\bm{\Theta}: \rho\,  U_\delta(\theta')\leq  y\} \geq
1-\bar\alpha\quad\Rightarrow\quad\pi_\delta \{\theta'\in\bm{\Theta}:
\rho\, U_\delta(\theta') >  y\}\leq \bar\alpha\, .
$$
Moreover,
$$
y < \rho\, y_{\bar\alpha}\quad\Rightarrow\quad\pi_\delta
\{\theta'\in\bm{\Theta}: \rho\,  U_\delta(\theta')\leq y\} <
1-\bar\alpha\quad\Rightarrow\quad \pi_\delta
\{\theta'\in\bm{\Theta}: \rho\,  U_\delta(\theta') > y\} >
\bar\alpha
$$
and taking the contrapositive one obtains
$$
\pi_\delta  \{\theta'\in\bm{\Theta}: \rho\, U_\delta(\theta') >
y\}\leq\bar\alpha\quad\Rightarrow\quad y\geq \rho\, y_{\bar\alpha}.
$$
Therefore $\{ \theta\in\bm{\Theta}: U_\delta(\theta) \geq\rho\,
y_{\bar\alpha} \}\equiv \{\theta\in\bm{\Theta}:
\pi_\delta\{\theta'\in\bm{\Theta}:\rho\,  U_\delta(\theta')>
U_\delta(\theta) \}\leq \bar\alpha \}$. We now derive a lower bound
on $\pi^{(J)}\{ \theta\in\bm{\Theta}:  U_\delta(\theta)\geq \rho\,
y_{\bar\alpha} \}$. Let us introduce the notation
$A_{\bar\alpha}:=\{\theta\in\bm{\Theta}:
U_\delta(\theta)<y_{\bar\alpha}\}$, $\bar
A_{\bar\alpha}:=\{\theta\in\bm{\Theta}:  U_\delta(\theta)\geq
y_{\bar\alpha}\}$, $B_{\bar\alpha,\rho}:=\{\theta\in\bm{\Theta}:
U_\delta(\theta)<\rho\, y_{\bar\alpha}\}$ and $\bar
B_{\bar\alpha,\rho}:=\{\theta\in\bm{\Theta}: U_\delta(\theta)\geq
\rho\, y_{\bar\alpha}\}$. Notice that $B_{\bar\alpha,\rho}\subseteq
A_{\bar\alpha}$ and $\bar A_{\bar\alpha} \subseteq \bar
B_{\bar\alpha,\rho}$. The quantity $\pi_\delta
\{\theta\in\bm{\Theta}:  U_\delta(\theta)< y\}$ as a function of $y$
is the left-continuous version of $\pi_\delta
\{\theta\in\bm{\Theta}:  U_\delta(\theta)\leq
y\}$\cite[p.162]{Gnedenko-68}. Hence, the definition of
$y_{\bar\alpha}$ implies $\pi_\delta(A_{\bar\alpha})\leq
1-\bar\alpha$ and $\pi_\delta(\bar A_{\bar\alpha})\geq\bar\alpha$.
Notice that
\begin{align*}
\pi_\delta(A_{\bar\alpha})\leq 1-\bar\alpha \quad&\Rightarrow\quad
\frac{\delta\pi_{Leb}(A_{\bar\alpha})} {\left[\int_{\bm{\Theta}}
U_\delta(\theta)\pi_{Leb}(d\theta)\right]} \leq
1-\bar\alpha\, ,\\
\pi_\delta(\bar A_{\bar\alpha}) \geq \bar\alpha \quad
&\Rightarrow\quad \frac{(1+\delta)\pi_{Leb}(\bar
A_{\bar\alpha})}{\left[\int_{\bm{\Theta}}
U_\delta(\theta)\pi_{Leb}(d\theta)\right]}\geq\bar\alpha\, .
\end{align*}
Hence, ${\pi_{Leb}(\bar A_{\bar\alpha})}>0$ and
$$
\frac{\pi_{Leb}(A_{\bar\alpha})}{\pi_{Leb}(\bar A_{\bar\alpha})}\leq
\frac{1-\bar\alpha}{\bar\alpha}\frac{1+\delta}{\delta}\, .
$$
Notice that ${\pi_{Leb}(\bar A_{\bar\alpha})}>0$ implies
${\pi_{Leb}(\bar B_{\bar\alpha,\rho})}>0$. We obtain
\begin{align*}
&\pi^{(J)} \{ \theta\in\bm{\Theta}:  U_\delta(\theta) \geq \rho\,
y_{\bar\alpha}\} =\frac{1} {\mbox{$\displaystyle 1 + \frac{\int_{
B_{\bar\alpha,\rho}} U_\delta(\theta)^J\pi_{Leb}(d\theta)}
{\int_{\bar
B_{\bar\alpha,\rho}}  U_\delta(\theta)^J\pi_{Leb}(d\theta)} $}}\\
&\geq\frac{1} {\mbox{$\displaystyle 1 +
\frac{\int_{B_{\bar\alpha,\rho}}
U_\delta(\theta)^J\pi_{Leb}(d\theta)}{\int_{\bar A_{\bar\alpha}}
U_\delta(\theta)^J\pi_{Leb}(d\theta)} $}} \geq \frac{1}
{\mbox{$\displaystyle 1 +
 \frac{\rho^{\, J}y_{\bar\alpha}^J}{y_{\bar\alpha}^J}\frac{\pi_{Leb}(B_{\bar\alpha,\rho})}{\pi_{Leb}(\bar A_{\bar\alpha})} $}}\\
&\geq \frac{1} {\mbox{$\displaystyle 1 +
\rho^{\,J}\frac{\pi_{Leb}(A_{\bar\alpha})}{\pi_{Leb}(\bar
A_{\bar\alpha})} $}} \geq \frac{1} {\mbox{$\displaystyle 1 +
\rho^{\,J}\frac{1-\bar\alpha}{\bar\alpha}\frac{1+\delta}{\delta}$}}\,
.
\end{align*}
Since $\{ \theta\in\bm{\Theta}:  U_\delta(\theta) \geq \rho\,
y_{\bar\alpha} \}\equiv \{\theta\in\bm{\Theta}:
\pi_\delta\{\theta'\in\bm{\Theta}: \rho\, U_\delta(\theta')>
U_\delta(\theta) \}\leq\bar\alpha \}$ the first part of the proof is
complete.

In the second part of the proof we show that the set
$\{\theta\in\bm{\Theta}: \pi_\delta\{\theta'\in\bm{\Theta}: \rho\,
U_\delta(\theta')> U_\delta(\theta) \}\leq\bar\alpha \}$ is
contained in the set of approximate global optimizers of $U$ with
value  imprecision $\tilde\epsilon := (\rho^{-1}-1)(1+\delta)$ and
residual domain $\tilde\alpha:=\frac{1+\delta}{\tilde\epsilon +
\delta}\,\bar\alpha$. Hence, we show that $\{\theta\in\bm{\Theta}:
\pi_\delta\{\theta'\in\bm{\Theta}: \rho\, U_\delta(\theta')>
U_\delta(\theta) \}\leq\bar\alpha \} \subseteq
\{\theta\in\bm{\Theta}: \pi_{Leb}\{\theta'\in\bm{\Theta}:
U(\theta')> U(\theta) +
\tilde\epsilon\}\leq\tilde\alpha\,\pi_{Leb}(\bm{\Theta}) \}.$ We
have
$$
U(\theta') > U(\theta)  + \tilde\epsilon\quad\Leftrightarrow \quad
\rho\, U_\delta(\theta') > \rho\, [U_\delta(\theta) +
\tilde\epsilon] \quad\Rightarrow\quad \rho\,
U_\delta(\theta')>U_\delta(\theta)
$$
which is proven by noticing that $\rho\,[U_\delta(\theta) +
\tilde\epsilon]\geq U_\delta(\theta)\,\, \Leftrightarrow\,\,  1-\rho
\geq U(\theta) (1-\rho)$\linebreak and $U(\theta)\in[0,\,1]$. Hence
$ \{\theta'\in\bm{\Theta}: \rho\, U_\delta(\theta')>
U_\delta(\theta) \} \,\supseteq \,
 \{\theta'\in\bm{\Theta}: U(\theta')> U(\theta) + \tilde\epsilon\} \, .$
Therefore $ \pi_\delta \{\theta'\in\bm{\Theta}: \rho\,
U_\delta(\theta')> U_\delta(\theta) \} \leq
\bar\alpha\,\Rightarrow\,\pi_\delta \{\theta'\in\bm{\Theta}:
U(\theta')> U(\theta) + \tilde\epsilon \} \leq \bar\alpha\, .$ Let
$Q_{\theta,\tilde\epsilon}:=  \{\theta'\in\bm{\Theta}: U(\theta')>
U(\theta) + \tilde\epsilon \}$ and notice that
$$
\pi_\delta\{\theta'\in\bm{\Theta}: U(\theta')>
U(\theta)+\tilde\epsilon \}=
 \frac{ \mbox{$\displaystyle
\int_{Q_{\theta,\tilde\epsilon}}  U(\theta')\pi_{Leb}(d\theta')
+\delta \pi_{Leb}(Q_{\theta,\tilde\epsilon})$} } {
\mbox{$\displaystyle \int_{\bm{\Theta}}
U(\theta')\pi_{Leb}(d\theta') + \delta\pi_{Leb}(\bm{\Theta})$} }\, .
$$
We obtain
\begin{align*}
\pi_\delta\{\theta'\in\bm{\Theta}: U(\theta')> U(\theta) +
\tilde\epsilon \} \leq \bar\alpha
&\Rightarrow\,\tilde\epsilon\,\pi_{Leb}(Q_{\theta,\tilde\epsilon}) +
\delta\pi_{Leb}(Q_{\theta,\tilde\epsilon})
\leq \bar\alpha (1+\delta)\pi_{Leb}(\bm{\Theta})\\
&\Rightarrow\,\pi_{Leb} \{\theta'\in\bm{\Theta}: U(\theta')>
U(\theta) +\tilde\epsilon \}  \leq \tilde\alpha\,
\pi_{Leb}(\bm{\Theta})\, .
\end{align*}
Hence we can conclude that
$$
\pi_\delta \{\theta'\in\bm{\Theta}: \rho\, U_\delta(\theta')>
U_\delta(\theta) \}\leq \bar\alpha\,\,\Rightarrow\,\,\pi_{Leb}
\{\theta'\in\bm{\Theta}: U(\theta')> U(\theta) +\tilde\epsilon \}
\leq \tilde\alpha\, \pi_{Leb}(\bm{\Theta})
$$
and the second part of the proof is complete.

We have shown that given $\bar\alpha\in (0,\, 1]$, $\rho\in
(0,\,1]$,
 $\tilde\epsilon := (\rho^{-1}-1)(1+\delta)$,
$\tilde\alpha:=\frac{1+\delta}{\tilde\epsilon + \delta}\,\bar\alpha$
and
$$
\sigma:=\frac{1} {\mbox{$\displaystyle 1 +
\rho^{\,J}\frac{1-\bar\alpha}{\bar\alpha}\frac{1+\delta}{\delta}$}}
= \frac { 1 } { \mbox{$\displaystyle 1 + \left[\frac{1 +
\delta}{\tilde\epsilon+1+\delta}\right]^{\,J}
\left[\frac{1}{\tilde\alpha}\frac{1 +\delta}{\tilde\epsilon+\delta}
- 1\right] \frac{1+\delta}{\delta} $}}\, ,
$$
the statement ``$\bm{\theta}$ is an approximate global optimizer of
$U$ with value imprecision $\tilde\epsilon$ and residual domain
$\tilde\alpha$" holds with probability at least  $\sigma$. Notice
that $\tilde\epsilon\in [0,\, 1]$ and $\tilde\alpha\in(0,\, 1]$ are
linked through a bijective relation to
$\rho\in[\frac{1+\delta}{2+\delta},\, 1]$  and $\bar\alpha\in (0,\,
\frac{\tilde\epsilon+\delta}{1+\delta}]$. The statement of the
theorem is eventually obtained by expressing $\sigma$ as a function
of desired $\tilde\epsilon=\epsilon$ and $\tilde\alpha=\alpha$.
\hfill$\Box$

{\small
\bibliographystyle{unsrt}
\bibliography{PAPERbib}
}

\end{document}